\documentclass[letterpaper]{article} 
\usepackage{xcolor}
\usepackage{aiva}
\usepackage{times}  
\usepackage{helvet}  
\usepackage{courier}  
\usepackage[hyphens]{url}  
\usepackage{graphicx} 
\usepackage{natbib}  
\usepackage{caption} 
\usepackage{algorithm}
\usepackage{xspace}
\usepackage{algorithmic}
\usepackage{amsfonts}
\usepackage{amsmath}
\usepackage{booktabs}
\usepackage{multirow}
\usepackage{placeins}
\usepackage{float}
\usepackage{newfloat}
\usepackage{listings}

\newcommand{\ours}{\texttt{AIVA}\xspace}

\DeclareCaptionStyle{ruled}{labelfont=normalfont,labelsep=colon,strut=off} 
\floatstyle{ruled}
\newfloat{listing}{tb}{lst}{}
\floatname{listing}{Listing}
\title{\ours: An AI-based Virtual Companion for Emotion-aware Interaction}

\author{
    Chenxi Li\textsuperscript{\rm 1}\thanks{Corresponding author.}
}
\affiliations{
    \textsuperscript{\rm 1}Glasgow College, University of Electronic Science and Technology of China\\
    ChenxiLi\_lcx@outlook.com
}

\usepackage{bibentry}

\begin{document}

\maketitle

\begin{abstract}
Recent advances in Large Language Models (LLMs) have significantly improved natural language understanding and generation, enhancing Human-Computer Interaction (HCI). However, LLMs are limited to unimodal text processing and lack the ability to interpret emotional cues from non-verbal signals, hindering more immersive and empathetic interactions. This work explores integrating multimodal sentiment perception into LLMs to create emotion-aware agents. We propose \ours, an AI-based virtual companion that captures multimodal sentiment cues, enabling emotionally aligned and animated HCI. \ours introduces a Multimodal Sentiment Perception Network (MSPN) using a cross-modal fusion transformer and supervised contrastive learning to provide emotional cues. Additionally, we develop an emotion-aware prompt engineering strategy for generating empathetic responses and integrate a Text-to-Speech (TTS) system and animated avatar module for expressive interactions. \ours provides a framework for emotion-aware agents with applications in companion robotics, social care, mental health, and human-centered AI.
\end{abstract}

\section{Introduction}

In recent years, Large Language Models (LLMs) have achieved remarkable progress in natural language understanding and generation \cite{zhao2023survey,yang2024harnessing}. Trained on massive-scale textual data with billions of parameters, LLMs such as ChatGPT, LLaMA, and ChatGLM have demonstrated impressive capabilities in diverse language tasks, including question answering, commonsense reasoning, and open-ended dialogue \cite{wang2023knowledge, yi2024survey,allemang2024increasing}. With their strong generalization and in-context learning abilities, LLMs have shown great potential to serve as the core engine in intelligent agents and virtual assistants, enabling more natural, flexible, and human-like interactions in a wide range of real-world applications \cite{xu2024mental,haltaufderheide2024ethics, thirunavukarasu2023large}.

\begin{figure}[t]
\centering
\includegraphics[width=1.0\columnwidth]{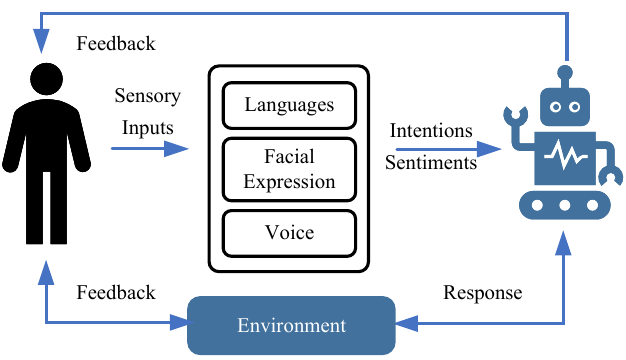}
\caption{An illustration of an ideal HCI system where user intentions and emotional states are perceived through multimodal signals, including language, facial expressions, and voice, to support empathetic and context-aware interaction.}
\label{fig:intro}
\vspace{-2.0pt}
\end{figure}

However, LLMs often struggle to accurately capture users’ emotional states, as emotional cues are not limited to textual inputs but are broadly expressed through non-verbal modalities, such as facial expressions, vocal tone, and gestures. For example, a simple sentence like ``It’s okay” may express completely different emotions depending on the user's facial expression or vocal tone, ranging from genuine acceptance to deep frustration or sadness. This limitation hinders the effective application of LLMs in the field of empathetic Human-Computer Interaction (HCI). An ideal HCI system, as illustrated in Figure~\ref{fig:intro}, perceives user intentions and affective states through multimodal cues, combining language, facial expressions, and voice to support emotionally intelligent interaction.

Multimodal affective computing aims to understand and model users' emotional states by analyzing signals from multiple modalities \cite{das2023multimodal,soleymani2017survey, poria2017review}. Recent advances in this field have led to the development of various architectures that integrate visual, acoustic, and textual information for sentiment classification, emotion recognition, and user state tracking \cite{du2022gated, zhang2022multimodal}. These systems have demonstrated strong performance in detecting fine-grained affective states across diverse scenarios \cite{peng2023fine,ye2022sentiment}. However, existing approaches are often developed in isolation from large language models (LLMs), limiting their applicability in interactive dialogue systems. Given the powerful in-context learning capabilities of LLMs, there is great potential to incorporate external multimodal sentiment perception modules into LLM-driven agents, enabling them to adapt their responses based on real-time emotional cues. 

Motivated by this, we propose \ours, an AI-based virtual agent that seamlessly integrates multimodal sentiment perception with large language models (LLMs) to enable emotion-aware and empathetic interactions. Specifically, \ours introduces a Multimodal Sentiment Perception Network (MSPN) that leverages a cross-modal fusion transformer and prototype-level supervised contrastive learning to extract users’ emotional states from both textual and visual cues. These emotion signals are then injected into the LLM via Emotion-aware Prompt Engineering (EPE), allowing it to generate context-sensitive and emotionally aligned responses. Finally, \ours incorporates a Text-to-Speech (TTS) system and an animated avatar module powered by Live2D, delivering expressive verbal and visual feedback to enable more human-like and engaging interactions.
Extensive experiments are conducted to validate both the sentiment recognition performance of MSPN and the empathetic interaction capability of the proposed \ours framework. The contributions of this work are summarized as follows:
\begin{itemize}
\item We propose a practical solution to bridge multimodal affective computing and LLMs, constructing an emotion-aware LLM-driven agent capable of perceiving multimodal sentiment cues and achieving empathetic interactions.
\item We introduce a novel multimodal sentiment analysis method, MSPN, which combines a cross-modal fusion transformer, sentiment prototypes, and prototype-level supervised contrastive learning to achieve fine-grained emotion recognition.
\item The proposed \ours framework demonstrates strong potential for real-world applications, such as virtual companions, social care agents, and emotionally intelligent human-computer interaction systems.
\end{itemize}

\section{Related Work}
\begin{figure*}[!t]
\centering
\includegraphics[width=1.0\textwidth]{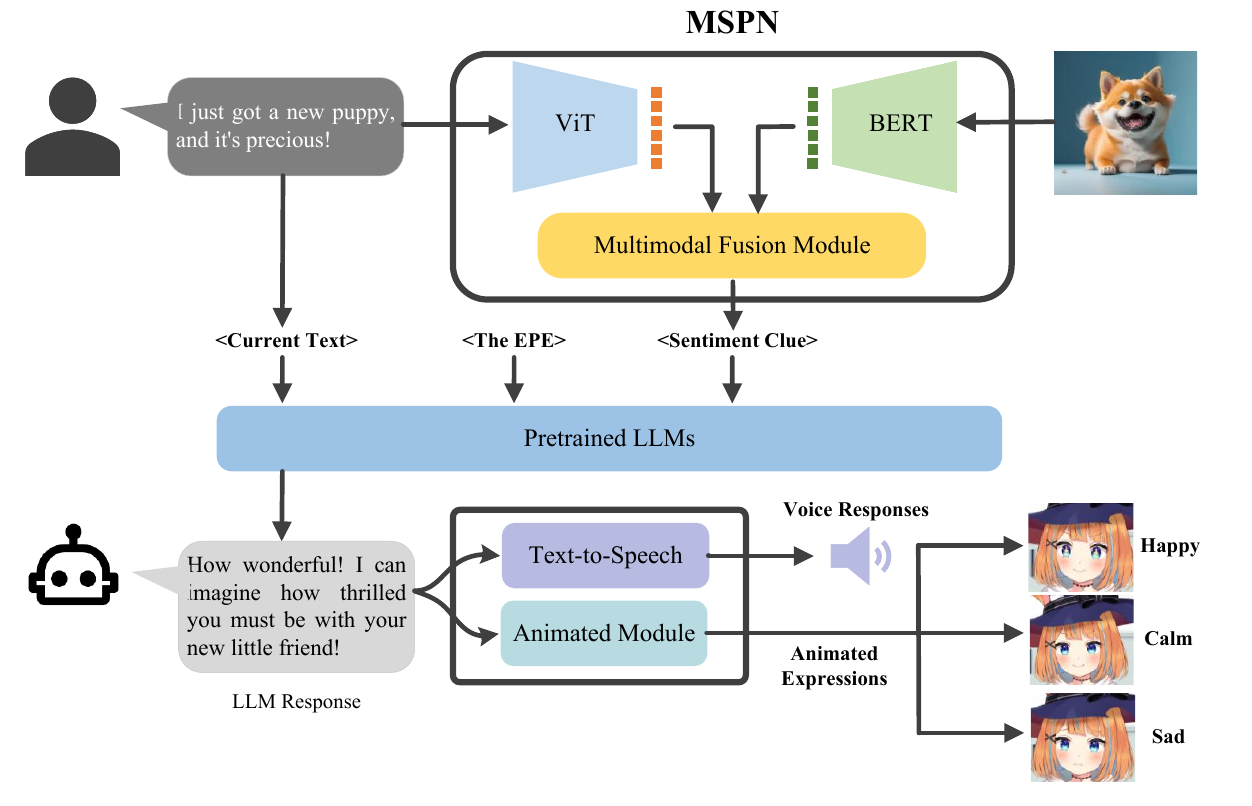}
\caption{The proposed \ours framework for empathetic HCI, integrating multimodal sentiment perception, LLM-driven empathetic responses, and expressive feedback through TTS and animated avatars.}
\label{fig:method_1}
\end{figure*}
\subsection{LLMs}
Large Language Models (LLMs) have made significant advancements in recent years, excelling in language understanding, commonsense reasoning, and question answering \cite{kumar2024large,bhat2023investigating}. These models are built on Transformer architectures, often containing billions of parameters, and are pre-trained on massive internet datasets to acquire world knowledge \cite{mckinzie2024mm1,patil2024review}. Notable LLMs like BERT \cite{devlin2019bert}, GPT \cite{radford2018improving}, and T5 \cite{raffel2020exploring}, each differing in architecture and pretraining methods. BERT uses a masked language model for contextual learning, while GPT employs autoregressive training for improved language generation. T5 integrates encoder-decoder architecture to handle diverse NLP tasks with a unified model. More recent models like LLaMA \cite{touvron2023llama} enhance training efficiency and performance, while instruction-finetuning approaches, as seen in Alpaca \cite{taori2023alpaca}, improve task-specific comprehension.  GLM \cite{du2021glm} further improves LLMs by combining generative pretraining and masked language modeling, offering a stronger performance in both language understanding and generation. Despite their success in textual tasks, LLMs face challenges in incorporating non-verbal cues, making them less effective for Human-Robot Interaction (HRI), where multi-modal data, including visual and speech signals are essential.

\subsection{Multimodal Learning}
Multimodal learning integrates data from multiple modalities (e.g., text, images, audio) to improve decision-making by providing comprehensive evidence \cite{ramachandram2017deep,xu2023multimodal}. This approach enhances neural networks' performance in cross-modal tasks by leveraging diverse data sources. Recent advancements in multimodal learning include CLIP \cite{radford2021learning} and ViLT \cite{kim2021vilt}, which utilize contrastive learning and transformers to align visual and textual representations. BLIP \cite{li2022blip} extended this by offering a unified approach for cross-modal tasks like classification and generation. More recent models, such as BLIP-2 \cite{li2023blip}, LLaVA \cite{liu2023visual}, and MiniGPT-4 \cite{zhu2023minigpt}, bridge the gap between vision and language, enabling LLMs to generate text based on visual input. These models employ innovative strategies like learnable projection layers and cross-attention mechanisms to enhance cross-modal alignment and improve reasoning capabilities in multimodal scenarios.

\section{Method}

\subsection{Overview of the proposed \ours}
The proposed \ours framework aims to integrate multimodal sentiment perception with LLMs to construct emotionally intelligent virtual companion, enabling empathetic HCI. As shown in Figure~\ref{fig:method_1}, \ours captures users’ emotional cues from both textual and visual inputs using the Multimodal Sentiment Perception Network (MSPN). MSPN fuses these modalities to extract sentiment signals, which are then used to guide the response generation of pre-trained LLMs through Emotion-aware Prompt Engineering (EPE). The framework is designed to generate contextually appropriate and empathetic language responses while providing expressive feedback through a Text-to-Speech (TTS) system and an animated avatar module. This seamless integration of multimodal perception and LLMs allows for more natural, engaging, and emotionally aligned interactions, providing a foundation for applications in virtual companions.

\subsection{The Proposed MSPN}
The proposed MSPN is designed to detect users' real-time emotional states and generate sentiment cues for LLMs. Given a batch of image-text pairs \( B = \{ (x_{1}, t_{1}), (x_{2}, t_{2}), \dots, (x_{N}, t_{N}) \} \), where \( N \) is the batch size, the Vision Transformer (ViT) is employed as the visual encoder to extract visual representations for each image. The visual representations are denoted as \( V_i = \{ v_1, v_2, \dots, v_M \} \), where \( v \in \mathbb{R}^{768} \) and \( M \) is the number of visual tokens. In parallel, the textual encoder (e.g., BERT) encodes the textual representations as \( T_i = \{ t_1, t_2, \dots, t_L \} \), where \( t \in \mathbb{R}^{768} \) and \( L \) denotes the length of textual tokens. These tokens are then processed by Cross Attention Fusion and the Cross-Modal Fusion Transformer.

\begin{figure*}[!t]
\centering
\includegraphics[width=1.0\textwidth]{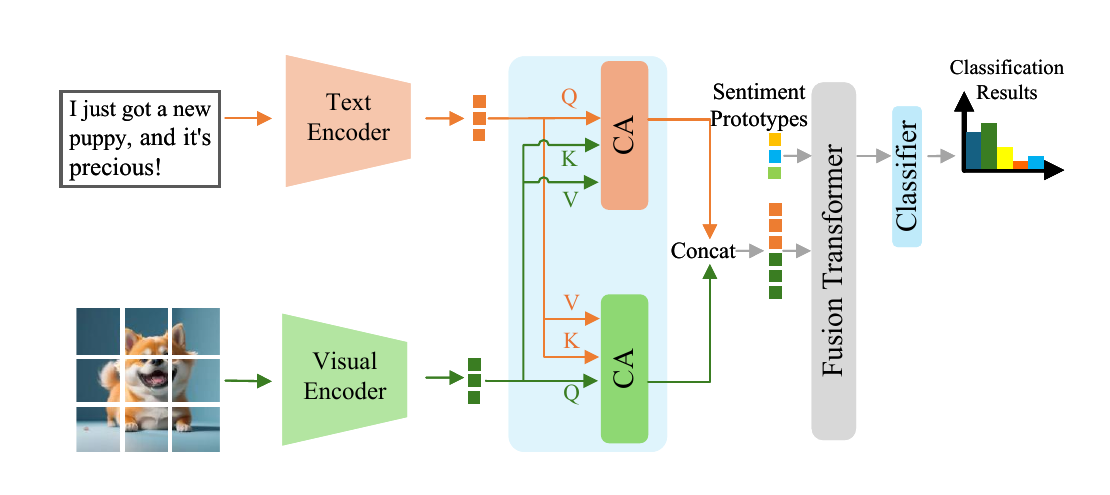}
\caption{The MSPN architecture, where visual and textual representations are fused using Cross-attention and cross-modal fusion transformer to give the final prediction results of user's current sentiment state.}
\label{fig:method_2}
\end{figure*}

\paragraph{Cross Attention Fusion}
To facilitate effective cross-modal interaction, the MSPN utilizes a Cross-Attention (CA) mechanism. For each modality, we use its corresponding representation as the query, while the other modality serves as both the key and value:
\begin{equation}
\left\{
\begin{aligned}
\hat{T}_i &= \text{softmax}\left(\frac{W_Q V_i (W_K T_i))^T}{\sqrt{d_k}}\right) W_V T_i \\
\hat{V}_i &= \text{softmax}\left(\frac{W_Q T_i (W_K V_i)^T}{\sqrt{d_k}}\right) W_V V_i
\end{aligned}
\right.
\end{equation}

The enhanced representations effectively capture the crucial cross-modal relationships and are used for subsequent modality fusion. Finally, \( \hat{T}_i \) and \( \hat{V}_i \) are combined to form a unified multimodal representation \( Z_i^0 \):
\begin{equation}
Z_i^0 = \left[ \hat{T}_i ; \hat{V}_i \right]
\end{equation}
Where \( Z_i^0 \in \mathbb{R}^{(L+M) \times 768} \) represents the joint multimodal features, integrating both image-related and text-related information into one comprehensive representation.

\begin{figure}[t]
\centering
\includegraphics[width=1.0\columnwidth]{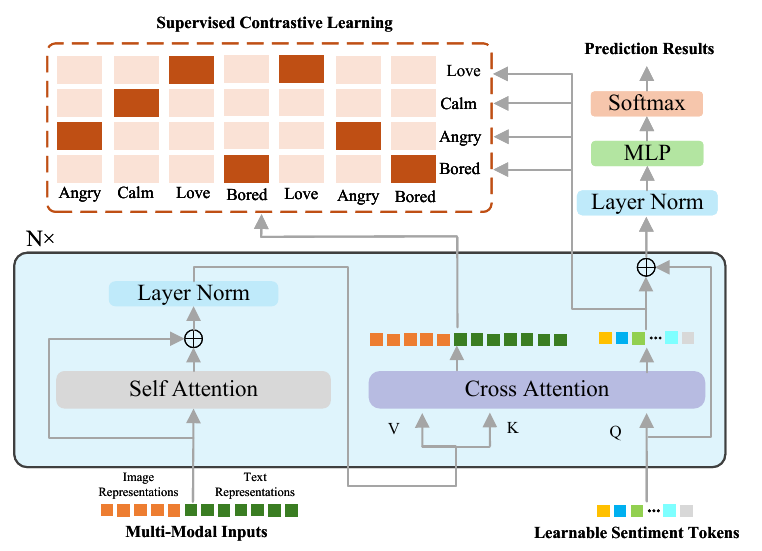} 
\caption{The designed cross-modal fusion transformer.}
\label{fig:method_3}
\end{figure}

\paragraph{Cross-Modal Fusion Transformer} 
The multimodal representation \( Z_i^0 \) is then sent to a cross-modal fusion transformer. This fusion transformer is designed to achieve deep fusion between visual and textual modalities, as shown in Figure~\ref{fig:method_3}. Here, we introduce a group of learnable sentiment prototypes, which can be denoted as:
\begin{equation}
S^0 = \{s_1, s_2, \dots, s_c\},\quad S \in \mathbb{R}^{C \times 768}
\end{equation}
Where $C$ denotes the number of sentiment category. These sentiment prototypes are designed to capture sentiment-specific information through CA mechanisms from the unified multimodal representations. Specifically, at the $j$-th transformer layer, the $S^{j}$ are computed as:
\begin{equation}
{S}^{j} = \text{softmax}\left(\frac{W_Q S^{j-1} (W_K Z_{i}^{j})^T}{\sqrt{d_k}}\right) W_V Z_{i}^{j} + S^{j-1}
\end{equation}
Where $Z_{i}^{j}$ represents the unified multimodal representation at the $j$-th transformer layer. The sentiment prototypes output by the final layer $S^N$ is then sent to the classifier $MLP(\cdot)$ for sentiment recognition.
\begin{equation}
\hat{p}_i = \text{softmax}(MLP(S^N))
\end{equation}

\paragraph{Training Objectives} 
To facilitate effective representation learning in the MSPN, we utilize both classification and supervised contrastive learning objectives. These objectives enable the model to predict the sentiment of multimodal inputs and align sentiment prototypes with corresponding multi-modal representations. Specifically, the classification loss is computed using cross-entropy with the ground-truth labels:
\begin{equation}
L_{\text{classification}} = - \sum_{i=1}^{n_{\text{batch}}} y_i \log(\hat{p}_i)
\end{equation}
Where \( y_i \) represents the ground-truth label in one-hot vector format. In addition, the supervised contrastive loss aligns the multi-modal representations \( Z_i^N \) with their corresponding sentiment prototypes \( S^N \) at the final layer. This loss is formulated as:
\begin{equation}
L^{\text{z} \rightarrow \text{s}}_{\text{sup}} = - \sum_{i \in B} \frac{1}{|P(i)|} \sum_{p \in P(i)} \log \left( \frac{\exp \left( f(z_i^N, s_p^N) \right)}{\sum\limits_{a \in A(i)} \exp \left( f(z_i^N, s_a^N) \right)} \right)
\end{equation}
Where \( B \) is the batch of anchor samples, \( P(i) \) is the set of positive samples, and \( A(i) \) represents all samples in the batch except the anchor. The function \( f(\cdot, \cdot) \) is a similarity measure, typically computed as normalized cosine similarity. In the same way, prototype to multimodal representation alignment is also applied in the opposite direction, where each sentiment prototype serves as the anchor, and the corresponding multimodal representation is regarded as a positive sample.
\begin{equation}
L^{\text{s} \rightarrow \text{z}}_{\text{sup}} = - \sum_{j \in C} \frac{1}{|P(j)|} \sum_{p \in P(j)} \log \left( \frac{\exp \left( f(s_j^N, z_p^N)  \right)}{\sum\limits_{a \in A(j)} \exp \left( f(s_j^N, z_a^N) \right)} \right)
\end{equation}
Where \( C \) is the set of sentiment categories. The final training loss is the sum of the forward and reverse directions with classification loss:
\begin{equation}
L= L_{\text{classification}}+ \frac{\lambda}{2}(L^{\text{s} \rightarrow \text{z}}_{\text{sup}} + L^{\text{z} \rightarrow \text{s}}_{\text{sup}})
\end{equation}
Where $\lambda$ is a trade-off hyperparameter to control the balance between classification and contrastive learning loss.

\subsection{Sentiment-aware Prompt Engineering}
\begin{figure}[t]
\centering
\includegraphics[width=1.0\columnwidth]{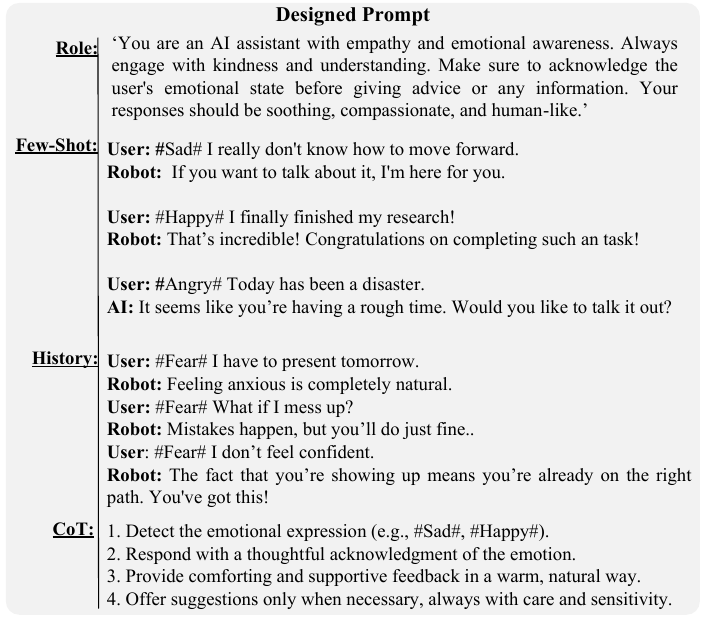}
\caption{The designed prompt template.}
\label{fig:method_4}
\end{figure}

To inject sentiment clues captured by the MSPN into the LLM, we use Emotion-aware Prompt Engineering (EPE). This approach adapts the sentiment classification results as a prefix in the prompt, guiding the LLM to generate emotionally aligned responses. By embedding sentiment information, the model is able to recognize the user's emotions and respond empathetically. Specifically, the designed prompt template includes key elements: role definition to ensure compassionate responses, few-shot examples to guide emotional responses, historical context to track emotional changes, and Chain-of-Thought (CoT) prompting to reason step by step before generating responses. This template is illustrated in Figure~\ref{fig:method_4}.

\subsection{Text-to-Speech and Animated Avatar}
To enhance the user experience and provide a more immersive and engaging interaction, we incorporate TTS and an Animated Avatar. The TTS system, powered by GPT-SoVITS \cite{gpt_sovits}, converts the language model's textual output into natural, emotionally expressive speech, ensuring that the system responses reflect the user's emotional state. To complement the speech, an animated avatar, created using Live2D, provides dynamic facial expressions and body movements, further enhancing the emotional engagement of the interaction. This combination of TTS and Animated Avatar allows for a more natural, human-like interaction, making the human-computer communication feel more intuitive and empathetic. 

Thus, \ours constructs an empathetic companion capable of understanding and responding to users' emotional states. By integrating MSPN with pre-trained LLMs, EPE, TTS, and Animated Avatar, \ours provides natural, contextually appropriate, and emotionally aligned interactions.

\begin{table}[h]
\centering
\caption{The Results on the MVSA datasets}
\resizebox{\columnwidth}{!}{  
\begin{tabular}{cccccc}
\hline
\multirow{2}{*}{} & \multirow{2}{*}{\textbf{Method}} & \multicolumn{2}{c}{\textbf{MVSA-Single}} & \multicolumn{2}{c}{\textbf{MVSA-Multi}} \\ \cline{3-6} 
                            &              & Accuracy $\uparrow$ & F1-Score $\uparrow$ & Accuracy $\uparrow$ & F1-Score $\uparrow$ \\
\hline
\multirow{3}{*}{Text}       & CNN          & 68.19\%        & 55.90\%       & 65.64\%        & 57.66\%       \\
                            & BiLSTM       & 70.12\%        & 65.06\%       & 67.90\%        & 67.90\%       \\
                            & BERT         & 71.11\%        & 69.70\%       & 67.59\%        & 66.24\%       \\
\hline
\multirow{2}{*}{Image}      & ResNet-50    & 64.67\%        & 61.55\%       & 61.88\%        & 60.98\%       \\
                            & OSDA         & 66.75\%        & 66.51\%       & 66.62\%        & 66.23\%       \\
\hline
\multirow{5}{*}{Image-Text} & MultiSentNet & 69.84\%        & 69.84\%       & 68.86\%        & 68.11\%       \\
                            & HSAN         & 69.88\%        & 66.90\%       & 67.96\%        & 67.76\%       \\
                            & Co-MN-Hop6   & 70.51\%        & 70.01\%       & 68.92\%        & 68.83\%       \\
                            & MGNNS       & 73.77\%        & 72.70\%       & 72.49\%        & 69.34\%       \\
                     & \textbf{MSPN}                  & \textbf{74.25\%}    & \textbf{72.84\%}   & \textbf{73.48\%}   & \textbf{70.01\%}   \\
\hline
\end{tabular}
}  
\label{tab:comparison_results}
\end{table}
\begin{table}[h]
\caption{The Results of the MSPN on TumEmo}
\centering
\resizebox{\columnwidth}{!}{
\begin{tabular}{ccccc}
\hline
 & Accuracy $\uparrow$ & Precision $\uparrow$ & Recall $\uparrow$ & F1-Score $\uparrow$ \\ \cline{2-5} 
\textbf{MSPN}  & 81.81\%                 & 82.48\%                       & 81.81\%                    & 81.89\%                \\
\hline
\end{tabular}}
\label{tab:pretraining_results}
\end{table}
\begin{figure}[t]
\centering
\includegraphics[width=1.0\columnwidth]{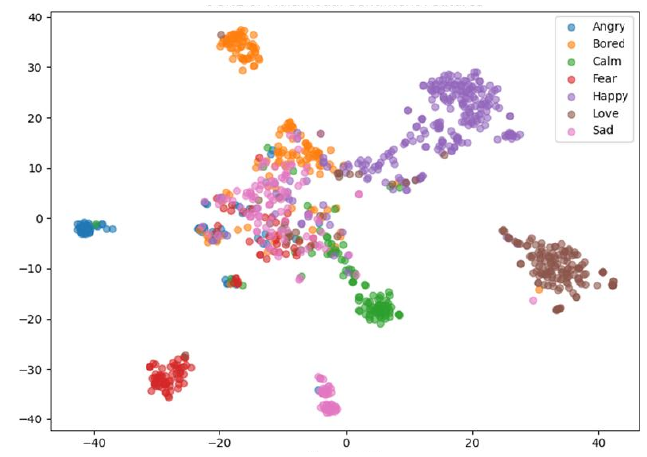}
\caption{The t-SNE results of the learned sentiment prototypes on the TumEmo datasets.}
\label{fig:pretraining_results}
\end{figure}
\begin{figure}[t]
\centering
\includegraphics[width=1.0\columnwidth]{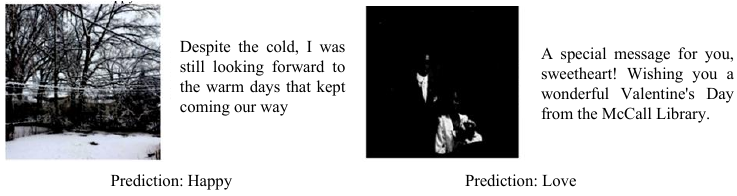}
\caption{The qualitative examples of the MSPN.}
\label{fig:example_pretraining}
\end{figure}

\section{Experiment}
\subsection{Setup}
\subsubsection{Datasets}
we evaluate the proposed MSPN on on several multimodal sentiment analysis datasets. Specifically, MVSA-Single \cite{niu2016sentiment} is a benchmark dataset with images and text labeled as positive, neutral, or negative, enabling fine-tuning for sentiment classification. MVSA-Multiple \cite{niu2016sentiment} extends this with multi-label sentiment annotations, providing a more robust evaluation of the model’s ability to handle nuanced emotional states across modalities. Moreover, TumEmo \cite{yang2020image} consists of over 195,000 image-text-emotion triplets, used for pre-training the MSPN to learn cross-modal sentiment signals.

\subsubsection{Models and Baselines} 
We compare our MSPN with several baselines, including OSDA \cite{yang2020image}, MultiSentNet \cite{xu2017multisentinet}, HSAN \cite{xu2017analyzing}, Co-MN-Hop6 \cite{xu2018co}, and MGNNS \cite{yang2021multimodal}. For \ours, we select the LLaMA2-Chat \cite{touvron2023llama} as the pretrained LLM.

\subsubsection{Hyperparameters Setting} For all experiments, we set the learning rate to \( 2 \times 10^{-5} \), and the batch size is set to 16 for evaluation on the two MVSA datasets and 24 for pretraining on the TumEmo dataset. The Adam optimizer is employed with standard momentum values. Training is conducted for one epoch on the TumEmo dataset for large-scale pretraining, and for 10 epochs on each of the MVSA datasets.

\subsection{Results}

\paragraph{Comparison Results on MVSA} Table~\ref{tab:comparison_results} presents the sentiment classification results across several models on the MVSA-Single and MVSA-Multi datasets. Text-based models, particularly BERT, outperform image-based models like ResNet-50 and OSDA, with BERT achieving the highest accuracy and F1 scores in the text modality. Multimodal models, such as MGNNS, show improved performance over unimodal models, with MSPN outperforming all baselines, achieving 74.25\% accuracy and 72.84\% F1 on MVSA-Single and 73.48\% accuracy and 70.01\% F1 on MVSA-Multi. These results demonstrate that MSPN effectively combines multimodal sentiment perception, significantly improving sentiment classification performance.

\paragraph{Pretraining Results on TumEmo}
Table~\ref{tab:pretraining_results} presents the pretraining results of MSPN on the TumEmo dataset. The model achieves 81.81\% accuracy, 82.48\% precision, 81.81\% recall, and 81.89\% F1 score, demonstrating its strong performance in multimodal sentiment recognition. These results indicate that the MSPN effectively learns sentiment signals from both visual and textual modalities, providing a solid foundation for further fine-tuning. Additionally, the t-SNE visualization of the learned sentiment prototypes in Figure~\ref{fig:pretraining_results} shows the model’s ability to effectively separate sentiment categories in the feature space. The distinct clustering of emotions like Happy, Sad and Angry demonstrates that \ours captures meaningful differences between emotional states. Some overlap, such as between Fear and Sad, indicates shared emotional characteristics, but overall, the model distinguishes between emotions well, highlighting its effectiveness in sentiment classification.

\paragraph{Qualitative Results on TumEmo}
Figure~\ref{fig:example_pretraining} shows qualitative examples of MSPN. The model accurately predicts emotions based on both textual and visual cues, demonstrating its ability to capture nuanced sentiment. For instance, the model correctly associates a winter landscape with a Happy sentiment, while it identifies a sad emotional state from an image of a crying character. Similarly, the Love sentiment is linked to an image of a couple, and the Angry sentiment is reflected in an intense scene, showing that MSPN can effectively process multimodal inputs for sentiment recognition.

\begin{figure*}[!t]
\centering
\includegraphics[width=1.0\textwidth]{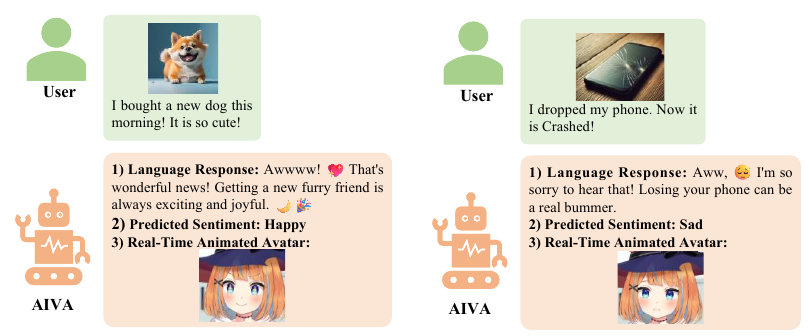}
\caption{The qualitative examples of the \ours.}
\label{fig:example_aiva}
\end{figure*}

\paragraph{Qualitative Results of \ours}
Figure~\ref{fig:example_aiva} presents qualitative examples of \ours in action. The system successfully responds empathetically to user emotions by generating contextually appropriate language and animated avatars in real-time. In the first example, when the user shares the excitement of getting a new dog, \ours responds with a cheerful message and a happy avatar. In the second example, when the user expresses frustration over dropping their phone, \ours provides a sympathetic response with a sad sentiment and a corresponding avatar. These examples demonstrate how \ours effectively combines language generation, sentiment prediction, and real-time animated feedback to create emotionally aware interactions.

\subsection{Ablation Study}

\paragraph{Ablation of the MSPN}
Here, we conduct the ablation experiments to verify the effectiveness of different components in MSPN, including Cross Attention Fusion (CAF), Cross-Modal Fusion Transformer (CMFT), and Supervised Contrastive Learning (SCL). The results, shown in Table~\ref{tab:ablation}, demonstrate that removing any component leads to a drop in performance, highlighting the importance of each part. Specifically, CAF plays a critical role in fusion quality, CMFT enhances multimodal feature alignment, and SCL ensures robust sentiment classification. The full model MSPN, achieves the best performance, outperforming all ablated variants.

\begin{table}[t]
\centering
\caption{The ablation results of the MSPN}
\resizebox{\columnwidth}{!}{
\begin{tabular}{ccccc}
\hline
\multirow{2}{*}{\textbf{Model}} & \multicolumn{2}{c}{\textbf{MVSA-Single}}         & \multicolumn{2}{c}{\textbf{MVSA-Multi}}          \\ \cline{2-5} 
                                & Accuracy $\uparrow$ & F1-Score $\uparrow$ & Accuracy $\uparrow$ & F1-Score $\uparrow$ \\
\hline
w/o CAF  & 72.94\% & 71.17\% & 18.70\% & 67.41\% \\
w/o CMFT & 71.68\% & 71.19\% & 70.76\% & 68.76\% \\
w/o SCL  & 73.01\% & 71.78\% & 69.41\% & 68.25\% \\
\textbf{MSPN}                   & \textbf{74.25\%}        & \textbf{72.84\%}       & \textbf{73.48\%}        & \textbf{70.01\%}       \\
\hline
\end{tabular}}
\label{tab:ablation}
\end{table}
\begin{table}[t]
\centering
\caption{The ablation results on the $\lambda$}
\resizebox{\columnwidth}{!}{
\begin{tabular}{ccccc}
\hline
\multirow{2}{*}{\textbf{$\lambda$}} & \multicolumn{2}{c}{\textbf{MVSA-Single}} & \multicolumn{2}{c}{\textbf{MVSA-Multi}} \\ \cline{2-5} 
    & Accuracy $\uparrow$ & F1-Score $\uparrow$ & Accuracy $\uparrow$ & F1-Score $\uparrow$ \\
\hline
0.5 & 73.45\%        & 72.17\%       & 71.23\%        & 68.32\%       \\
1.0 & 74.25\%        & 72.84\%       & 73.48\%        & 70.01\%       \\
1.5 & 73.45\%        & 72.61\%       & 69.64\%        & 67.21\%       \\
2.0 & 72.12\%        & 70.43\%       & 70.05\%        & 68.36\%       \\
\hline
\end{tabular}}
\label{tab:lambda}
\end{table}

\subsubsection{Evaluation on the Sensitivity of $\lambda$}
Table~\ref{tab:lambda} presents the results of the ablation study on the hyperparameter $\lambda$. The results show that the model performs best when $\lambda = 1.0$, achieving 74.25\% accuracy and 72.84\% F1 score on the MVSA-Single dataset, and 73.48\% accuracy and 70.01\% F1 score on the MVSA-Multi dataset. Lower values of $\lambda$ (0.5) result in reduced performance, while higher values (1.5 and 2.0) also cause a drop in performance, especially in terms of F1 score. This suggests that an appropriate balance between the classification and contrastive loss is crucial for optimal performance.

\section{Conclusion}
In this work, we proposed \ours, a framework that integrates multimodal sentiment perception with LLMs to enable emotionally intelligent human-robot interactions. By combining the Multimodal Sentiment Perception Network (MSPN), Emotion-aware Prompt Engineering (EPE), and expressive output through Text-to-Speech (TTS) and Animated Avatar, \ours enhances the emotional engagement and naturalness of interactions. Our extensive experiments demonstrate that \ours outperforms existing methods in sentiment classification, emotion alignment, and multimodal interaction, providing a solid foundation for virtual companions and other human-centered AI applications.

\bibliography{aiva}

\end{document}